\documentclass[twocolumn]{article}

\usepackage[T1]{fontenc}
\usepackage{mathptmx}           
\usepackage[numbers,square]{natbib}
\usepackage{amsmath,amssymb,amsfonts}
\usepackage{bm}
\usepackage{algorithmic}
\usepackage{graphicx}
\usepackage{url}
\usepackage{hyperref}
\usepackage{textcomp}
\usepackage{placeins}
\usepackage{subcaption}
\usepackage{caption}
\usepackage{xcolor}
\usepackage{enumitem}

\hypersetup{
  colorlinks=true,
  linkcolor=purple,
  urlcolor=blue,
  citecolor=cyan,
  anchorcolor=black,
  pdfborder={0 0 0},
  breaklinks=true
}

\def\BibTeX{{\rm B\kern-.05em{\sc i\kern-.025em b}\kern-.08em
    T\kern-.1667em\lower.7ex\hbox{E}\kern-.125emX}}

\begin{document}

\title{Histopathological Spectrum-Guided Prostate Stratification via Segmentation-Assisted Diagnostic Transformer}

\author{
  \small
  Leyang Li$^{1}$, Lihua Chen$^{2}$, Huangang Hu$^{3}$, Tianhang Hao$^{2}$,\\
  \small
  Hao Cheng$^{4}$, Xin Zhang$^{5}$, Qianru Sun$^{6}$,
  Bingxu Lu$^{1,\dagger}$,\\
  \small
  Wenlong Yu$^{6,\dagger}$, and Feng Duan$^{1}$\\[4pt]
  \small
  $^{1}$College of Artificial Intelligence, Nankai University,
  Tianjin 300350, China\\
  \small
  $^{2}$Tianjin First Central Hospital, Tianjin 300382, China\\
  \small
  $^{3}$School of Electronics and Information Engineering,
  Tiangong University, Tianjin 300387, China\\
  \small
  $^{4}$School of Artificial Intelligence,
  Hebei University of Technology, Tianjin 300401, China\\
  \small
  $^{5}$North China Digital Health Technology Co., Ltd., Jinan 250117, China\\
  \small
  $^{6}$School of Computing and Information Systems,
  Singapore Management University, Singapore 178902, Singapore\\[4pt]
  \small
  $^{\dagger}$Corresponding authors: Bingxu Lu
  (lubingxu@nankai.edu.cn) and Wenlong Yu
  (wenlongyu@smu.edu.sg).
}

\maketitle

\begingroup
\renewcommand{\thefootnote}{}
\footnotetext{This work has been submitted to the IEEE for possible publication. Copyright may be transferred without notice, after which this version may no longer be accessible.}
\endgroup

\begin{abstract}
Prostate cancer diagnosis with multiparametric MRI (mpMRI) is commonly based on PI-RADS assessment or binary classification, which suffer from subjectivity and fail to capture clinically relevant pathological heterogeneity. To address this limitation, we construct a Prostate Cancer Histopathology Spectrum Dataset (PCa-HSD) and formulate a clinically meaningful four-class classification task, addressing the underrepresentation of benign lesions that are easily confounded with prostate cancer in existing datasets. We propose Language-guided Segmentation-assisted Diagnostic Transformer model (LSDT), which leverages zero-shot segmentation to provide anatomical priors and performs effective multi-modal slice fusion for classification. Our proposed method consistently improves accuracy across backbones, achieving the best average accuracy of 0.633 and JointRecall of 0.768 in five-fold cross-validation on a cohort of 344 patients. These results demonstrate that integrating pathology supervision and anatomical priors significantly enhances fine-grained prostate MRI classification and provides a more clinically relevant paradigm for risk stratification. Code will be made publicly available in a future revision.
\end{abstract}


\section{Introduction}
\label{sec:introduction}
Prostate cancer (PCa) is the second most common malignancy among men worldwide~\cite{rawla2019epidemiology}, and it exhibits pronounced heterogeneity in both clinical presentation and pathological progression~\cite{shoag2016clinical}. As a key non-invasive imaging tool, mpMRI has been widely adopted for PCa detection and risk stratification. Combined with the Prostate Imaging Reporting and Data System (PI-RADS)~\cite{weinreb2016pi}, mpMRI can effectively identify clinically significant cancer, reduce unnecessary biopsy procedures, and help avoid overtreatment of indolent lesions, thereby serving as a cornerstone of contemporary imaging-based assessment~\cite{cheng2023avoiding}.

\begin{figure}[!t]
\begin{minipage}{0.95\linewidth}
\centering
\includegraphics[width=\linewidth]{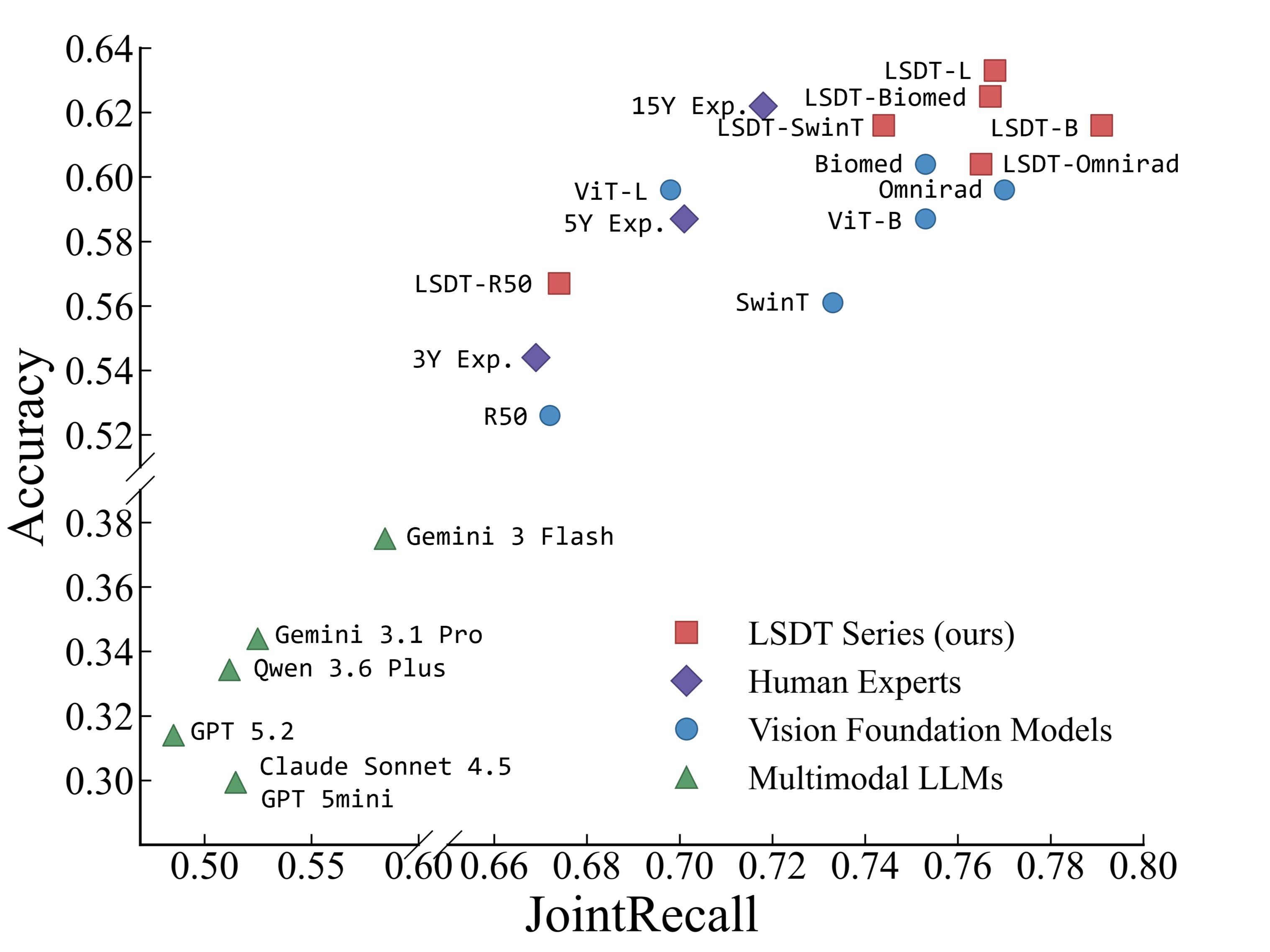}
\captionsetup{font=footnotesize}
\captionof{figure}{
Performance comparison on the internal PCa-HSD cohort across human experts, vision foundation models, prompt-based MLLMs, and the proposed LSDT variants.LSDT-Large achieves the best overall performance, with average ACC of 0.633 and JointRecall of 0.768.
}
\label{fig.abstract_performance}
\end{minipage}
\end{figure}

However, current mpMRI-based PCa analysis still faces limitations in several key aspects, which hinder precise alignment with the pathological gold standard. In clinical decision-making, the PI-RADS score is inherently a subjective interpretation system based on imaging features, and it inevitably deviates from histopathology~\cite{drevik2022utility,esengur2025multimodal,palmisano2024analysis,franz2025biopsy}. Prior studies have shown that, even under standardized protocols, PI-RADS exhibits substantial inter-reader variability for intermediate-risk lesions (e.g., PI-RADS 3--4)~\cite{greer2019interreader}. Moreover, its mapping to pathological status is imperfect: only 15-20\% of PI-RADS 3 lesions are confirmed to harbor clinically significant prostate cancer (sPCa)~\cite{esengur2025multimodal,palmisano2024analysis}. These observations suggest that PI-RADS alone is insufficient for accurately predicting true histological risk.

Most existing studies focus on binary classification (e.g., clinically significant vs.\ non-significant)~\cite{rajagopal2024mixed,yang2024deep,castillo2021classification,cao2019joint,juneja2025prostate,horasan2024advancing,litjens2014computer,wang2018automated,devente2021deep}, or directly use PI-RADS scores as training labels~\cite{yildirim2022deep,rossi2020multi}. This paradigm not only inherits PI-RADS subjectivity but also overlooks fine-grained pathological information that is critical for clinical decision-making. In practice, lesion management depends not merely on benign/malignant status but also on specific pathological subtypes: clinically significant cancer (commonly defined as ISUP grade $\geq 2$ or Gleason score $\geq 7$) is more aggressive and typically requires active intervention, whereas clinically non-significant cancer (ISUP grade 1 or Gleason score $=6$) may be suitable for active surveillance~\cite{epstein20162014,heidenreich2008eau}. Meanwhile, benign prostatic diseases are highly prevalent in the aging male population, among which benign prostatic hyperplasia (BPH) remains the most prevalent benign prostate disorder requiring chronic medical management~\cite{zurowska2023prostate}. Moreover, BPH represents one of the most common conditions and exhibits imaging characteristics similar to prostate cancer, making the differentiation between benign lesions, clinically non-significant cancer and clinically significant cancer challenging~\cite{purysko2021pitfalls}. Simplifying the problem to binary classification therefore provides insufficient evidence for precise risk stratification~\cite{yu2025fine}.

Despite significant progress in deep learning for prostate MRI interpretation~\cite{rajagopal2024mixed,yang2024deep,castillo2021classification,cao2019joint,hung2022cat,yildirim2022deep,rossi2020multi,lu2018feature}, several methodological challenges remain. Many approaches underutilize the inherent three-dimensional spatial information in mpMRI, often relying on 2D or pseudo-3D strategies that discard spatial context~\cite{hung2024cross}. Meanwhile, effectively fusing complementary information across multi-modal sequences such as T2-weighted imaging (T2WI), apparent diffusion coefficient (ADC), and diffusion-weighted imaging (DWI) remains technically challenging.

To address these limitations, this paper constructs a Prostate Cancer Histopathology Spectrum Dataset, termed \textbf{PCa-HSD}, which enables fine-grained classification beyond conventional settings, providing a more objective and clinically meaningful reference standard. Moreover, we introduce a framework that is both clinically grounded in pathology and technically designed to exploit inter-slice sequential features from multi-modal MRI. Specifically, we define four categories: normal prostate, benign prostatic hyperplasia (BPH), clinically non-significant prostate cancer (nsPCa), and clinically significant prostate cancer (sPCa). This fine-grained categorization better reflects real-world clinical risk stratification, where management decisions depend not only on the presence of cancer but also on its pathological aggressiveness and differentiation from benign conditions. To effectively tackle the challenges posed by the sequential nature of multi-modal data and the strong overlap in mpMRI appearance between BPH and both sPCa and nsPCa, we further design a Language-guided Segmentation-assisted Diagnostic Transformer model (LSDT) that first leverages the zero-shot segmentation capability of Segment Anything Model 3 (SAM3) to focus learning on the prostate region~\cite{carion2025sam}, and then integrates slice sequential features from T2WI, ADC, and DWI through an effective multi-modal fusion classifier for accurate discrimination among all four categories.

In summary, the main contributions of this work are as follows:
\begin{itemize}
  \item We construct a pathology-grounded prostate mpMRI dataset and benchmark for four-class classification, uniquely annotated with biopsy-confirmed labels, providing a higher clinical fidelity resource and enabling more meaningful imaging--pathology research.
  \item We propose a Segmentation-assisted tri-modal slice-sequential classification network that incorporates prior knowledge from a foundation model SAM3 and integrates slice sequential features across slices from T2WI, ADC, and DWI to improve performance on the fine-grained task.
  \item We demonstrate strong performance on both the proposed pathology-grounded benchmark and the public PI-CAI dataset, providing cross-institutional generalization and quantitative evidence suggesting potential clinical utility for supporting more precise risk stratification and decision-making.
\end{itemize}

\section{Related Works}

\subsection{Tasks on Prostate mpMRI}
Existing studies still leave room for improvement in task definition and technical approaches. At the task level, most work focuses on binary classification (e.g., csPCa vs.\ non-csPCa or benign vs.\ malignant). For instance, Horasan and G\"une\c{s} (2024) proposed a 3D CNN-based framework for MRI-based prostate cancer detection~\cite{horasan2024advancing}, and Juneja \emph{et al.} (2024) developed PC-Net for automatic binary prostate cancer classification on MRI~\cite{juneja2025prostate}. However, binary settings overlook richer pathological categories commonly encountered in clinical practice (e.g., hyperplasia, normal tissue, and clinically non-significant cancer). Clinically significant cancer and non-significant cancer differ fundamentally in biological behavior and management strategies~\cite{heidenreich2008eau}; conflating them may lead to inappropriate treatment decisions. Therefore, finer-grained classification (e.g., normal, hyperplasia, non-significant cancer, significant cancer) is needed for more precise risk stratification.

\subsection{Multi-Modal Representation and Cross-Slice Fusion in mpMRI}
Existing methods often underutilize the inter-slice sequential information embedded in multi-modal mpMRI sequences. Many adopt 2D or 2.5D strategies, which process each slice independently or only exploit local neighboring slices, thus failing to model the long-range sequential variations and global structural transitions along the prostate longitudinal axis~\cite{hung2024cross,arif2020clinically,sun2023deep}. Besides, the effectiveness of multi-modal fusion across T2-weighted imaging, diffusion-weighted imaging, apparent diffusion coefficient and extended multimodal settings remains to be further improved, as many fusion schemes struggle to capture cross-modal interactions~\cite{cai2024fully,zhu2022multimodal}.

\subsection{Vision Foundation Models in Medical Imaging}
Vision foundation models such as the Segment Anything Model (SAM) have demonstrated strong zero-shot capability in natural image segmentation~\cite{kirillov2023segment,ravi2024sam2,zou2023segment,li2024segment}, but their direct application to medical images is challenging due to domain shift. Subsequent works, including MedSAM/MedSAM2, substantially improve performance on medical image segmentation via fine-tuning and domain adaptation~\cite{ma2024segment,ma2025medsam2,mazurowski2023segment,huang2024segment,chen2024ma}. However, most existing studies focus on segmentation itself, and the use of SAM-like foundation models to assist classification remains relatively underexplored. Their potential to guide classification networks toward regions of interest and improve fine-grained classification accuracy warrants further investigation.

\section{Method}

\begin{figure*}[!t]
  \centering
  \includegraphics[width=\textwidth]{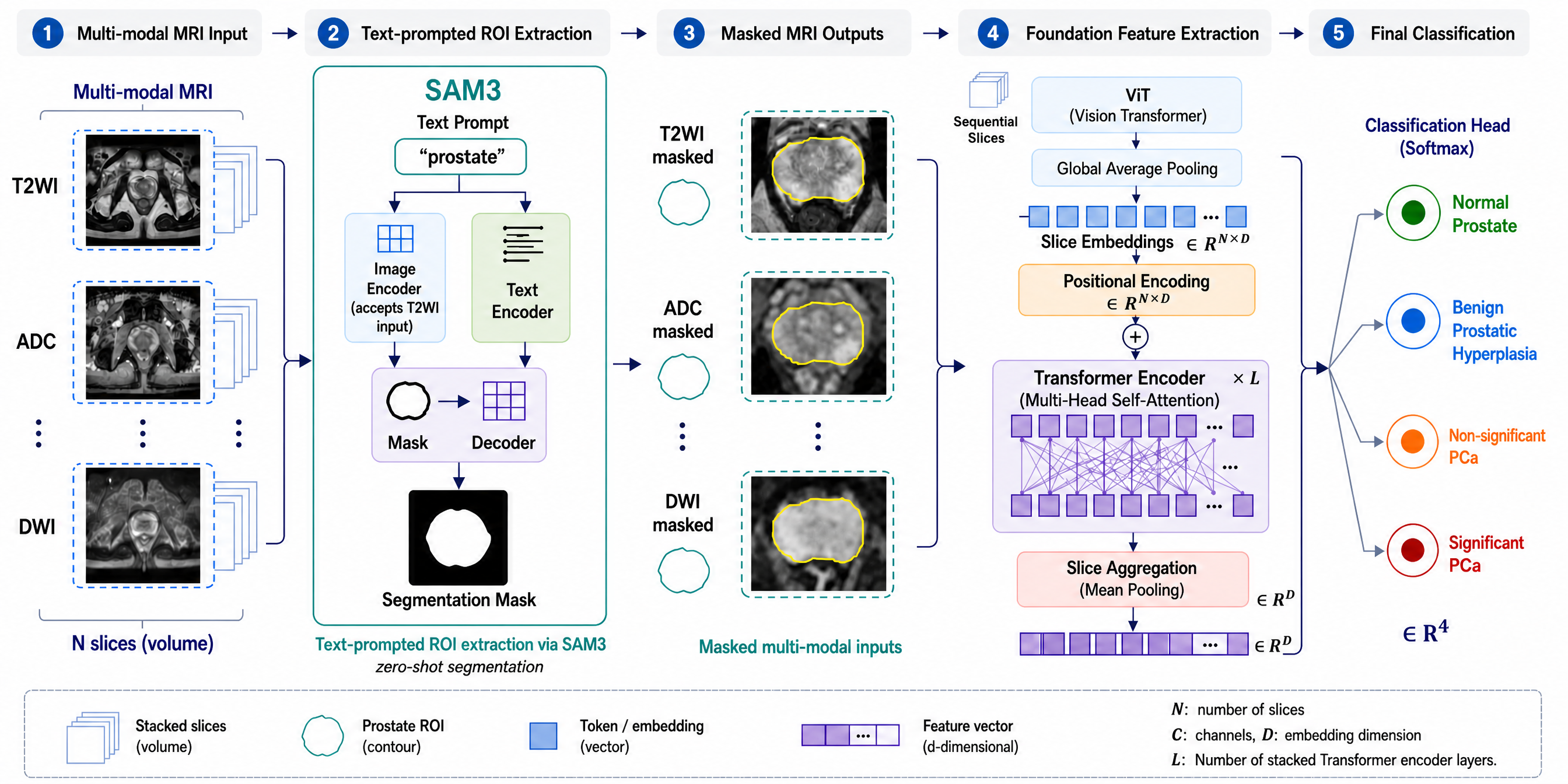}
  \caption{Overview of the proposed LSDT framework. Multi-modal MRI volumes (T2WI, ADC, DWI) are used as input. SAM3 processes the T2WI sequence with a text prompt ``prostate'' to generate a segmentation mask, which is then applied to all three modalities to extract the prostate region. The masked slices are fed into a pre-trained vision foundation model for feature extraction. Finally, slice-wise attention fusion aggregates the features across slices to produce the four-class prediction (normal, BPH, nsPCa, sPCa).}
  \label{fig.overview}
\end{figure*}

\subsection{Overview of the Proposed Framework}
As illustrated in Fig.~\ref{fig.overview}, we propose a multi-stage framework for automatic prostate status diagnosis using multi-modal MRI. The pipeline consists of three core components: (1) text-prompted region-of-interest (ROI) extraction, where SAM3 is used to separate the prostate region from background noise; (2) foundation-model-based feature extraction, where large-scale pre-trained models (e.g., ResNet and ViT) encode pathology-relevant patterns; and (3) slice-aware classification, where features across MRI slices are aggregated to classify the prostate status into four categories: normal prostate, benign prostatic hyperplasia (BPH), non-significant prostate cancer (nsPCa), and significant prostate cancer (sPCa).

\subsection{Multi-modal Input and Preprocessing}
The model takes multi-modal MRI sequences as input, including T2WI, ADC and DWI. These modalities provide complementary anatomical and functional information that is critical for accurate diagnosis~\cite{weinreb2016pi}. T2-weighted imaging offers high soft-tissue contrast and spatial resolution, allowing clear delineation of prostate zonal anatomy and lesion morphology; prostate cancers typically appear as focal hypointense regions relative to surrounding benign tissue on T2WI images. DWI measures the random Brownian motion of water molecules within tissue, with high b-value DWI highlighting areas of restricted diffusion that often correspond to increased cellular density in malignancy. The derived ADC map quantifies this diffusivity, and lower ADC values have been shown to correlate with higher Gleason grade tumors and more aggressive disease. The combination of these sequences in mpMRI has been demonstrated to improve diagnostic performance compared with any single modality alone, since T2WI reflects structural anatomy while DWI/ADC provide functional and microstructural cues. Before input to the model, images from each sequence are preprocessed to standardize orientation, resolution, and intensity, and co-registered to ensure voxel-wise alignment across modalities, enabling effective learning from their complementary information.

\subsection{Text-Prompted ROI Extraction via SAM}
To suppress irrelevant background tissues and focus the model on the prostate gland, we adopt a text-prompted ROI extraction strategy based on the latest SAM3, a foundation vision model capable of promptable concept segmentation. Unlike traditional segmentation models that require manual spatial prompts or task-specific annotations, SAM3 supports open-vocabulary segmentation using natural language prompts or exemplar images, enabling the model to localize and generate pixel-accurate masks for all instances matching a described concept without additional training on task-specific labels. This capability stems from its design to generalize to a broad set of visual concepts, allowing recognition and segmentation of previously unseen categories directly from text descriptions alone~\cite{carion2025sam}.

In our workflow, semantic natural language prompt ``prostate'' is used to drive SAM3 to automatically segment the prostate across multi-modal MRI slices. This zero-shot segmentation capability obviates the need for handcrafted spatial prompts or additional model fine-tuning, enabling efficient ROI extraction and cropping for T2WI, ADC, and DWI sequences. By focusing the input on anatomically relevant tissue, the extracted regions reduce background interference and provide more informative, high-resolution inputs to the downstream classification network, thereby improving the model's ability to learn discriminative features for fine-grained pathological classification.

\subsubsection{Mask Generation and Segmentation Inference}
For each MRI slice, the T2WI modality is replicated to three channels for alignment and then fed into the SAM3 image encoder to obtain visual features. The visual features, together with the text semantic embeddings generated from the prompt, are provided to the mask decoder. The mask decoder fuses image information with semantic prompts to predict the prostate segmentation and outputs an accurate binary mask.

\subsubsection{ROI Extraction and Cropping}
After obtaining the prostate mask for each slice, we perform post-processing and cropping. For each modality, the original MRI slice is multiplied pixel-wise by the corresponding binary prostate mask, producing a cropped \emph{masked image} in which the background is suppressed and only the prostate region is retained.

For modality $m\in\{\mathrm{T2WI},\mathrm{ADC},\mathrm{DWI}\}$ and slice index $s\in\{1,\ldots,S\}$, the masked slice is computed as
\begin{equation}
I_{\mathrm{masked}}^{(m,s)} = I^{(m,s)} \odot M^{(s)}, \forall s \in \{1,\dots,S\},
\end{equation}
where $I^{(m,s)}$ denotes the slice-$s$ image of modality $m$, $M^{(s)}$ is the corresponding binary mask, $\odot$ is the element-wise product, $I_{\mathrm{masked}}^{(m,s)}$ is the resulting masked image, and $S$ is the total number of slices for a patient.

\subsection{Feature Extraction Using Pre-trained Foundation Models}
We leverage vision foundation models pre-trained on large-scale datasets to fully exploit the benefits of transfer learning. By transferring robust feature representations learned from massive natural-image corpora, this strategy alleviates the limitations of scarce medical annotations and insufficient sample diversity. We employ pre-trained feature extractors represented by vision foundation models to process the masked multi-modal images. Specifically, T2WI, DWI, and ADC are stacked along the channel dimension and fed into the pre-trained visual model to obtain a high-dimensional visual embedding vector after global pooling.

For a given slice $s$, the multi-modal feature extraction process is expressed as
\begin{equation}
\mathbf{f} = \mathrm{Encoder}\Big(\mathrm{Concat}\big(I_{\mathrm{masked}}^{(\mathrm{T2WI},s)}, I_{\mathrm{masked}}^{(\mathrm{ADC},s)}, I_{\mathrm{masked}}^{(\mathrm{DWI},s)}\big)\Big),
\end{equation}
where $\mathrm{Concat}(\cdot)$ concatenates inputs along the channel dimension and $\mathbf{f}\in\mathbb{R}^{D}$ is the extracted $D$-dimensional feature vector. Since global pooling is used, the feature can be further expressed as $\mathbf{f}=\mathrm{GlobalAvgPool}(\mathrm{Encoder}(\cdot))$.

\subsection{Classification and Slice-Wise Feature Aggregation}
Let $\{\mathbf{f}_s\}_{s=1}^{S}$ denote slice-wise embeddings extracted by the foundation encoder for a patient with $S$ slices, with $\mathbf{f}_s\in\mathbb{R}^{D}$. To preserve anatomical ordering and enable long-range volumetric context modeling, we perform sequence modeling along the slice axis using learnable positional encoding and hierarchical self-attention.

Each slice token is first injected with positional information:
\begin{equation}
\tilde{\mathbf{f}}_s = \mathbf{f}_s + \mathbf{e}_s, \quad s=1,\ldots,S,
\end{equation}
where $\mathbf{e}_s\in\mathbb{R}^{D}$ is the learnable positional embedding.

Let $\mathbf{X}^{(0)}=[\tilde{\mathbf{f}}_1;\ldots;\tilde{\mathbf{f}}_S]\in\mathbb{R}^{S\times D}$. The sequence is then processed by a Transformer encoder consisting of $L$ stacked encoder layers. Each encoder layer comprises a multi-head self-attention (MSA) module followed by a feed-forward network, with residual connections and layer normalization applied after each sublayer. In the $l$-th encoder layer, the query, key, and value matrices are computed as
\begin{equation}
\mathbf{Q}=\mathbf{X}^{(l-1)}\mathbf{W}_Q,\
\mathbf{K}=\mathbf{X}^{(l-1)}\mathbf{W}_K,\
\mathbf{V}=\mathbf{X}^{(l-1)}\mathbf{W}_V,
\end{equation}
followed by
\begin{equation}
\mathrm{Attn}(\mathbf{Q},\mathbf{K},\mathbf{V})
=
\mathrm{Softmax}\!\left(
\frac{\mathbf{Q}\mathbf{K}^{\top}}{\sqrt{d}}
\right)\mathbf{V},
\end{equation}
where the attention weights quantify pairwise relevance between slice embeddings to perform adaptive cross-slice feature fusion.

For multi-head self-attention,
\begin{equation}
\mathrm{head}_i=\mathrm{Attn}(\mathbf{Q}_i,\mathbf{K}_i,\mathbf{V}_i),
\end{equation}
\begin{equation}
\mathrm{MSA}(\mathbf{X})
=
\mathrm{Concat}(\mathrm{head}_1,\ldots,\mathrm{head}_H)\mathbf{W}_O.
\end{equation}

After passing through all $L$ encoder layers, the resulting context-enhanced slice representations are denoted as $\mathbf{h}_{1:S}$.

Finally, the aggregated representation is obtained via mean pooling, followed by a lightweight prediction head to produce four-class logits $\mathbf{z}\in\mathbb{R}^{4}$ and probabilities $\mathbf{p}=\mathrm{Softmax}(\mathbf{z})$.

\section{Experiments}

\subsection{Experimental Setup}

\subsubsection{Datasets}
\textbf{PCa-HSD (internal dataset):} 
We retrospectively collected prostate mpMRI examinations from 3,810 patients who underwent MRI at Tianjin First Central Hospital and Tianjin Medical University Cancer Institute between January 2021 and August 2025. The inclusion criteria were: (1) no prior biopsy or any prostate-targeted treatment before MRI; and (2) biopsy or radical prostatectomy performed within 4 weeks after MRI with complete pathological reports. To ensure adequate imaging quality for downstream analysis, all examinations included in PCa-HSD satisfied a score of at least 4 in Prostate Imaging Quality (PI-QUAL) scoring system~\cite{giganti2020prostate}.

\begin{figure}[t!]
\centering
\begin{subfigure}[b]{0.47\linewidth}
    \centering
    \includegraphics[width=\linewidth]{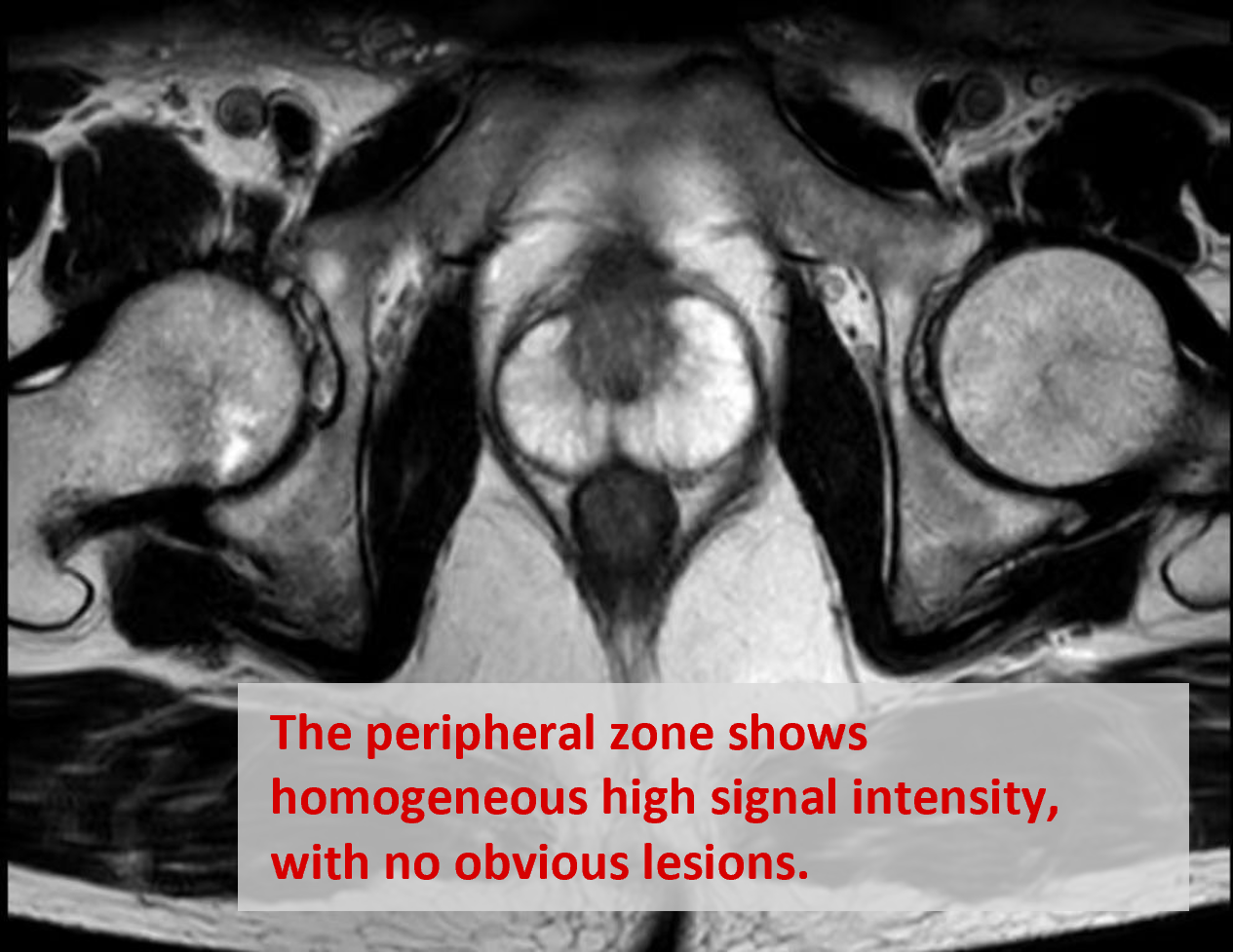}
    \caption{Normal}
    \label{fig.class_normal}
\end{subfigure}
\hfill
\begin{subfigure}[b]{0.47\linewidth}
    \centering
    \includegraphics[width=\linewidth]{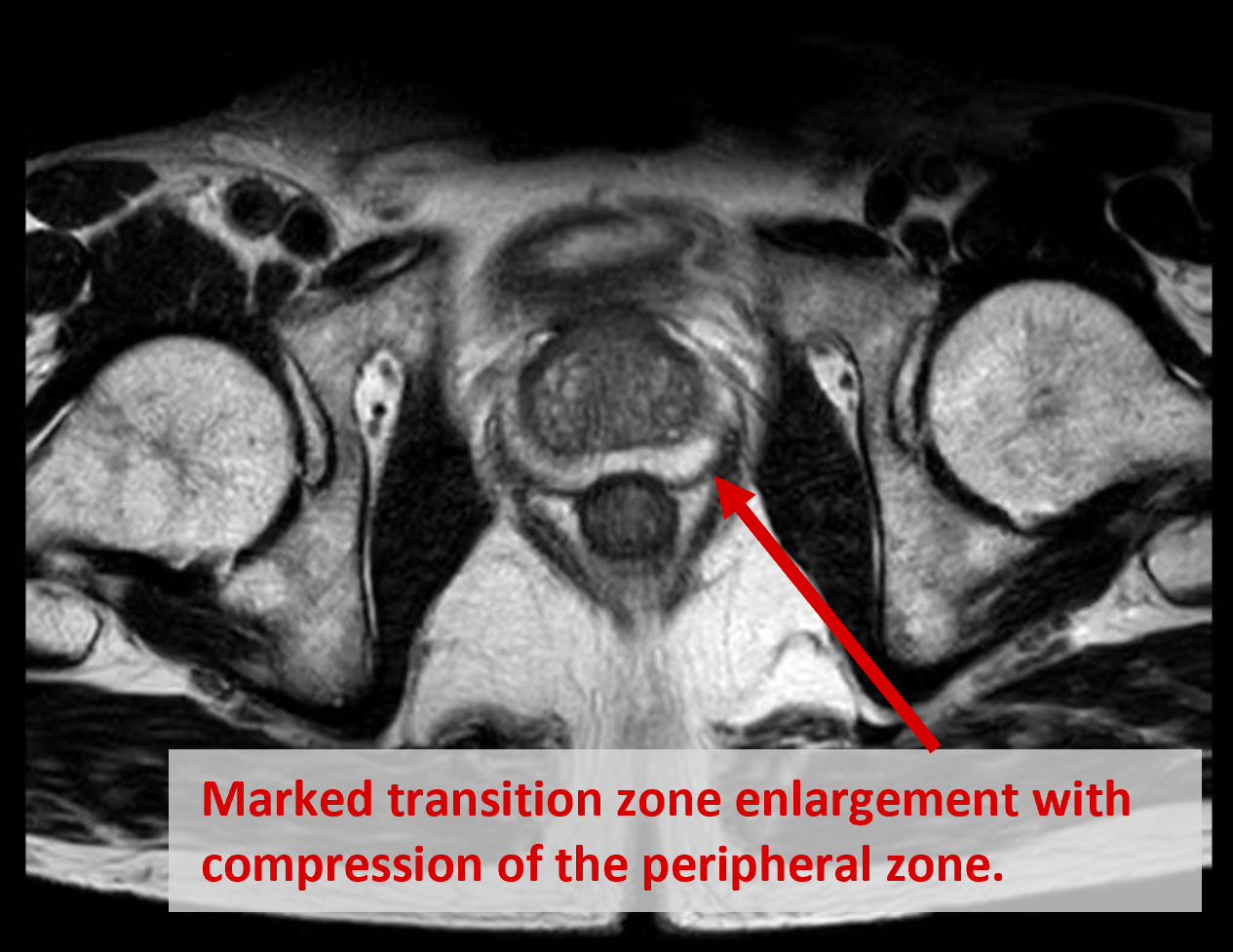}
    \caption{BPH}
    \label{fig.class_bph}
\end{subfigure}
\\[0.5em]
\begin{subfigure}[b]{0.47\linewidth}
    \centering
    \includegraphics[width=\linewidth]{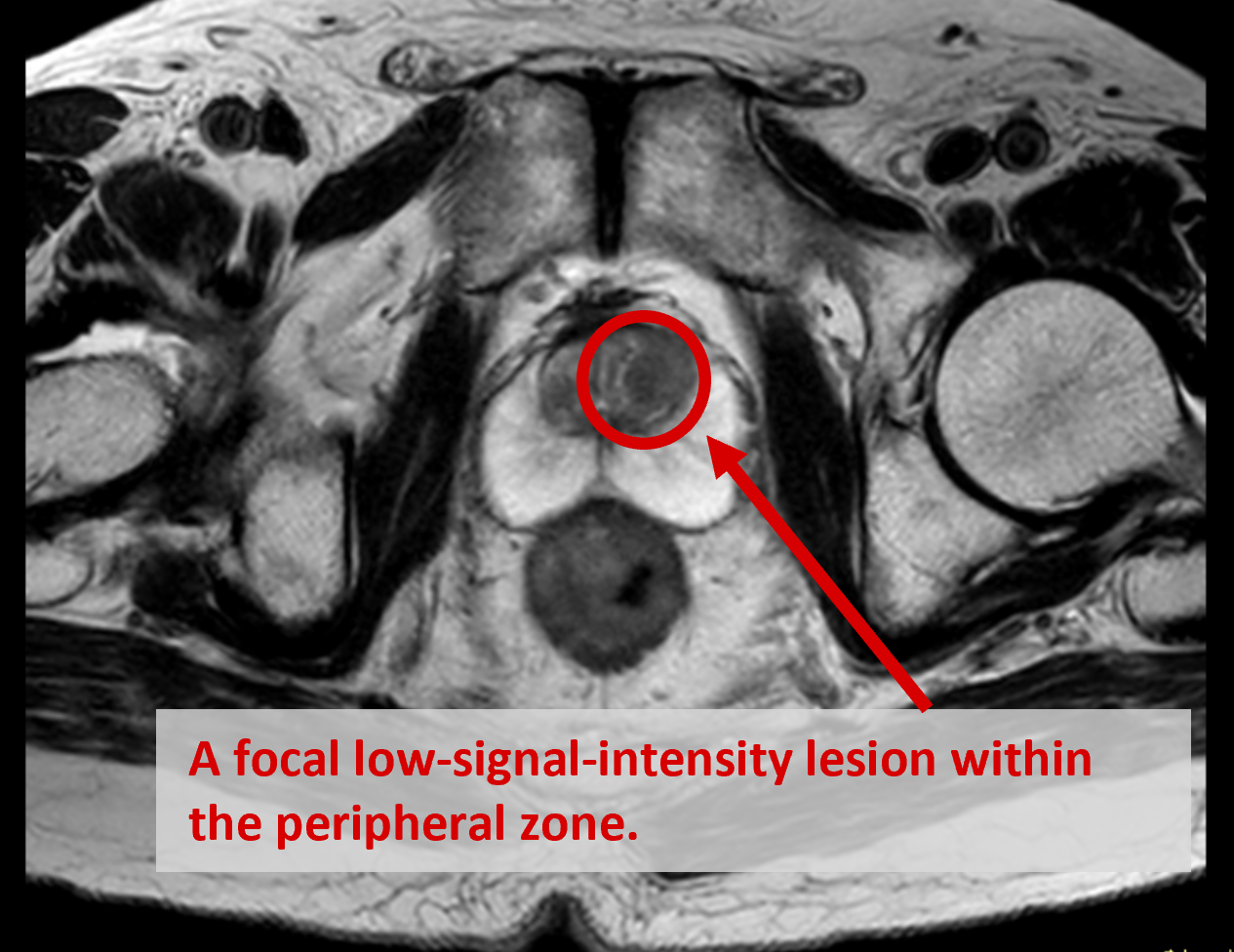}
    \caption{nsPCa}
    \label{fig.class_nspca}
\end{subfigure}
\hfill
\begin{subfigure}[b]{0.47\linewidth}
    \centering
    \includegraphics[width=\linewidth]{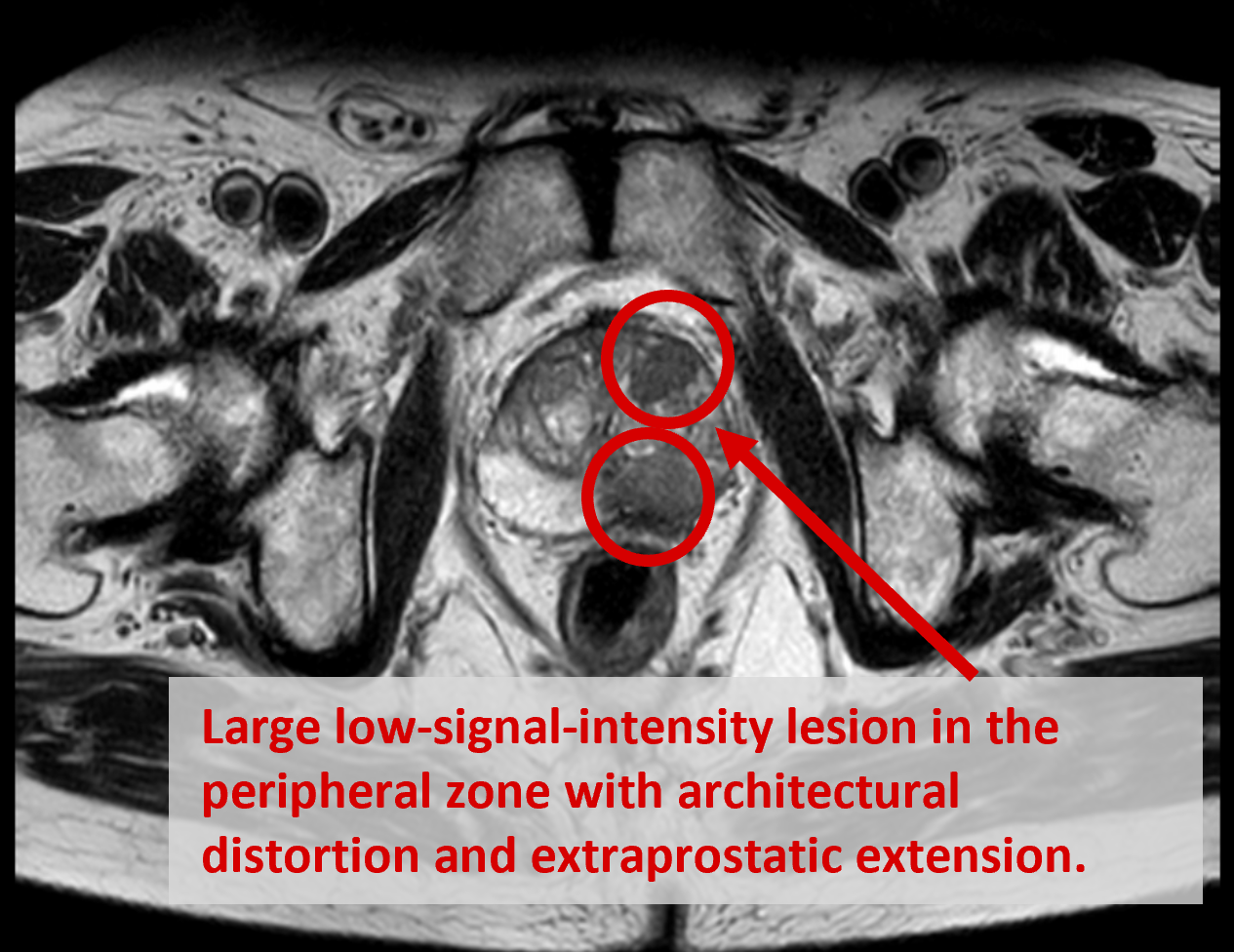}
    \caption{sPCa}
    \label{fig.class_spca}
\end{subfigure}
\caption{Representative T2WI examples of the four prostate tissue categories: (a) Normal; (b) BPH; (c) nsPCa; (d) sPCa.}
\label{fig.class_examples}
\end{figure}

Based on imaging findings and pathology, 344 patients were included in the final analysis and categorized into four groups: (1) normal prostate (Normal, n=68), subjects with no clinical evidence of prostate disease; (2) benign prostatic hyperplasia (BPH, n=99), confirmed as benign hyperplasia on pathology; (3) non-significant prostate cancer (nsPCa, n=76), prostate cancer with Gleason score $= 6$; and (4) significant prostate cancer (sPCa, n=101), prostate cancer with Gleason score $\geq 7$. Representative T2WI examples for each category are shown in Fig.~\ref{fig.class_examples}.

\textbf{External cohort (public binary dataset):} To evaluate cross-institutional generalization and mitigate potential institutional bias, we additionally adapted the public PI-CAI training and development cohort, which is a widely used multi-center benchmark for clinically significant prostate cancer (csPCa) detection~\cite{saha2023artificial}. This cohort contains 1,500 anonymized biparametric prostate MRI examinations from 1,476 patients collected at three Dutch centers between 2012 and 2021. The public release provides the core bpMRI sequences (T2-weighted and diffusion-based imaging, including DWI/ADC) and case-level reference targets for csPCa assessment defined under the PI-CAI protocol. To align this external cohort with our study design, we adapted PI-CAI into a patient-level binary classification task: cases labeled as csPCa were mapped to the positive class, and all remaining cases were mapped to non-csPCa. We then processed the PI-CAI cohort following the official PI-CAI data preparation and preprocessing protocol, and reported AUC as the primary external endpoint.

\subsubsection{Preprocessing and Data Augmentation}
All T2WI, ADC, and DWI volumes underwent N4ITK bias-field correction, intra-examination registration (axial T2WI as reference), and resampling to a unified spatial resolution. Subsequently, center cropping was applied to the central 16 slices covering the prostate region, all slices were resized to $224\times224$, and intensity values were min--max normalized to $[0,1]$. To improve robustness and mitigate overfitting, on-the-fly augmentation was performed during training, including random rotation within $[-15^\circ,15^\circ]$ and horizontal flipping (each with probability 0.5). Considering the prostate gland exhibits substantial variability in spatial location and size across slices and across patients in MRI, we applied a random affine transformation, consisting of translation within $\pm10\%$ of the image size and isotropic scaling in the range $[0.9, 1.1]$, while no shear was introduced in this operation.

\subsubsection{Implementation Details}
LSDT was built upon pretrained vision foundation models with a slice-fusion Transformer module, followed by a two-layer fully connected classifier with GELU activation to produce four-class output. For ROI extraction, a SAM3-based promptable segmentation model with frozen weights was used, where the text prompt was set to ``prostate'' and the generated masks were applied via element-wise multiplication. The slice-fusion Transformer module consisted of a two-layer transformer encoder, which processed slice-wise features to aggregate inter-slice information.

All models were implemented in PyTorch and trained on a single NVIDIA H20 GPU. Five-fold cross-validation was conducted at the patient level on the internal cohort. All backbone variants shared identical preprocessing, input resolution, data split, and optimization settings. The model was optimized using the Adam optimizer with a fixed learning rate of $1\times10^{-5}$ and trained for 100 epochs with a batch size of 16. Cross-entropy loss was used for supervision. Each fold required approximately 1.5 hours for training, and inference took 0.1--0.2 seconds per patient. For additional comparison on the internal PCa-HSD cohort, we further evaluated representative models from the GPT-5 series, Gemini 3 series, Claude Sonnet 4.5, and Qwen 3.6 Plus as multimodal large language model (MLLM) baselines under the same four-class task. Since these methods rely on prompt-driven multimodal inference rather than end-to-end supervised optimization within our framework, their results are reported separately from the backbone-based comparisons.

\subsubsection{Human Expert and MLLM Evaluation Protocol}
To provide a clinically grounded reference on the internal four-class cohort, we organized an independent reader study involving three radiologists with 3, 5, and 15 years of clinical experience, who independently reviewed all cases. Each reader independently reviewed the T2WI, ADC, and DWI examinations and assigned one patient-level label from the same four-class label space as PCa-HSD, namely Normal, BPH, nsPCa, or sPCa. The readers were blinded to the pathological reports and model predictions. All three readers independently reviewed the entire internal cohort at the patient level rather than a fold-wise subset.

For MLLM evaluation, all prompt-based models were tested on the same internal PCa-HSD four-class task without task-specific fine-tuning. Each model received a textual prompt containing the four-class task definition together with an explanation of the visual input format. As the image prompt, for each case we selected the 16 middle slices and constructed a long composite image by concatenating the three modalities (T2WI, ADC, and DWI) for each slice group. Each model was instructed to return a structured JSON object with three fields: predict label, predict confidence and key findings. Since these methods operate through prompt-driven multimodal inference rather than supervised optimization on PCa-HSD, their results are reported separately from the backbone-based and LSDT-based comparisons.

\subsubsection{Evaluation Metrics and Endpoints}
For the internal four-class cohort, we report Accuracy (ACC) and JointRecall as primary metrics. In screening and triage, two clinical demands are equally critical: correctly classifying benign patients to spare them from unnecessary biopsies, and correctly detecting cancer patients to ensure timely oncologic intervention. To jointly evaluate both capabilities, we merge Normal and BPH into a benign group and nsPCa and sPCa into a cancer group, then compute the proportion of correctly classified samples in this merged binary task. In this way, JointRecall simultaneously reflects the model's ability to reduce unnecessary biopsies for benign patients and avoid missing cancer patients who require further oncologic attention. JointRecall is computed as:
\begin{equation}
\mathrm{JointRecall}=\frac{TP_{\mathrm{benign}}+TP_{\mathrm{cancer}}}{N_{\mathrm{total}}},
\end{equation}
where $TP_{\mathrm{benign}}$ and $TP_{\mathrm{cancer}}$ denote the number of benign samples (Normal and BPH) correctly classified as benign and the number of cancer samples (nsPCa and sPCa) correctly classified as cancer, respectively, and $N_{\mathrm{total}}$ is the total number of samples.

For the public binary cohort, we report area under the ROC curve (AUC) as the main metric:
\begin{equation}
\mathrm{AUC}=\int_{0}^{1} \mathrm{TPR}(u)\,du,
\end{equation}
where $u=\mathrm{FPR}$ is the false-positive rate and $\mathrm{TPR}(u)$ is the corresponding true-positive rate on the ROC curve.

\begin{figure}[t!]
\centering
\begin{subfigure}[b]{0.49\linewidth}
    \includegraphics[width=\linewidth]{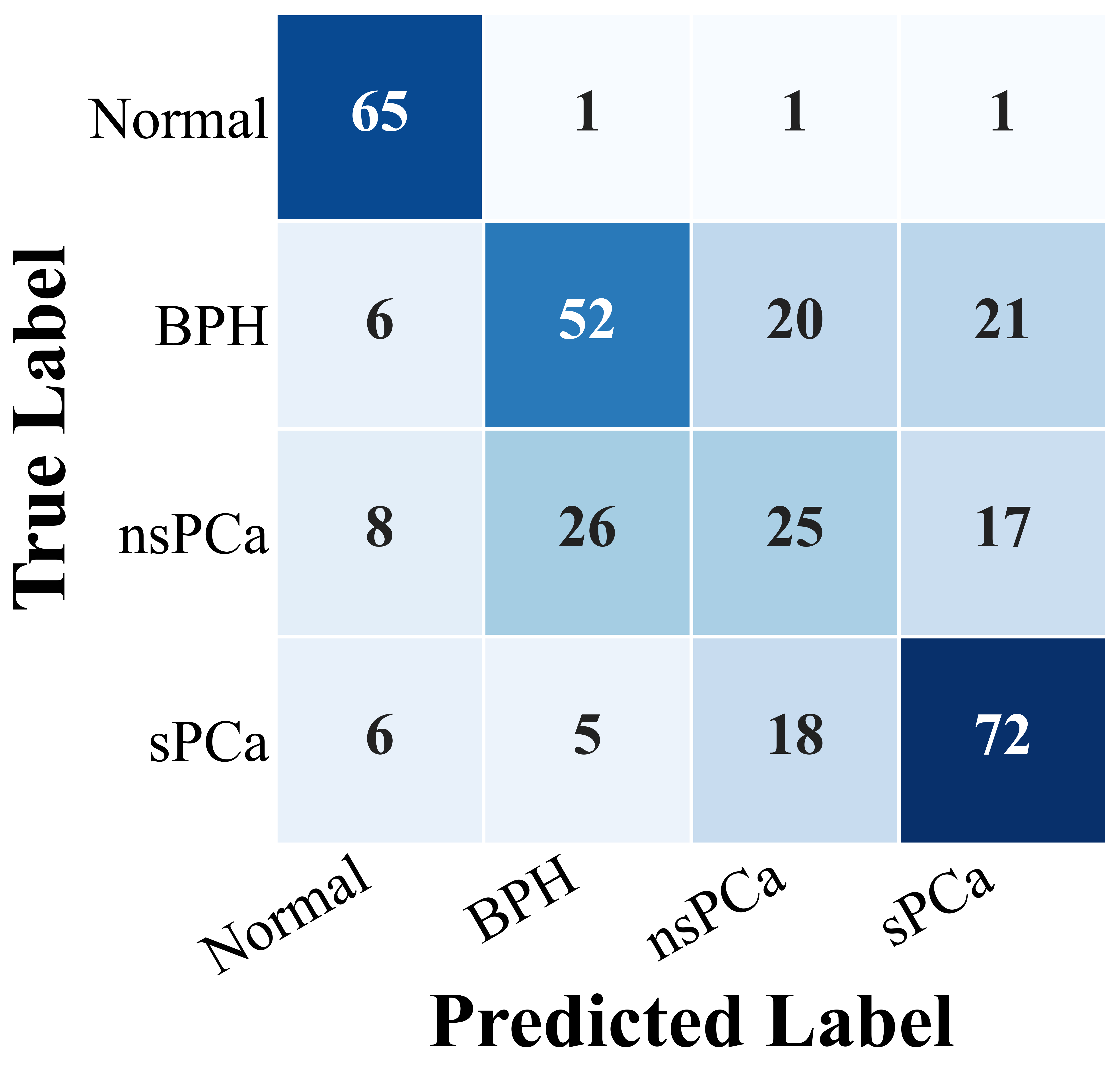}
    \caption{Senior radiologist}
\end{subfigure}
\hfill
\begin{subfigure}[b]{0.49\linewidth}
    \includegraphics[width=\linewidth]{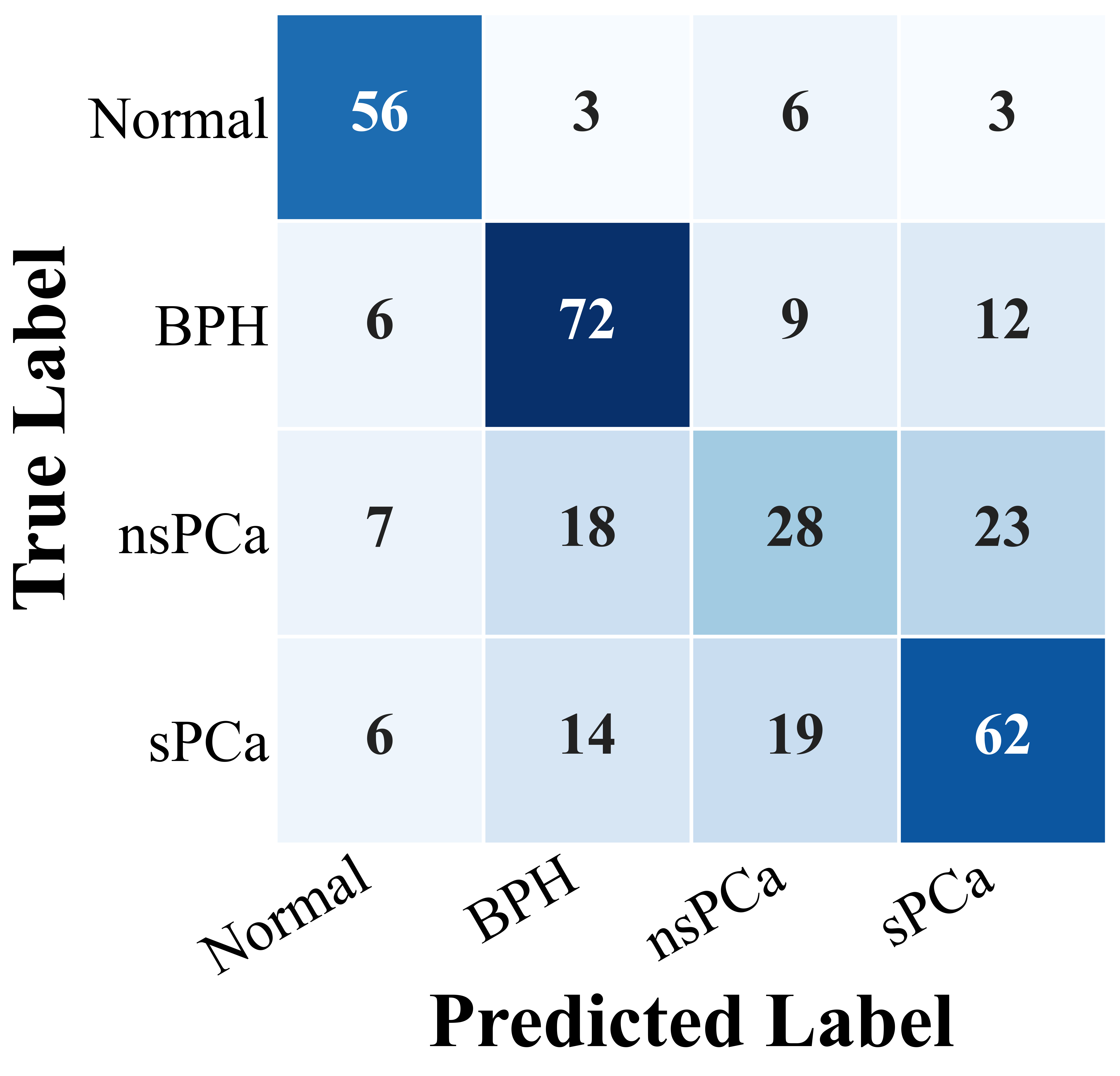}
    \caption{LSDT-Large}
\end{subfigure}
\caption{Confusion matrices on the internal PCa-HSD Four-Class Cohort.
(a) Performance of a senior radiologist with 15 years of experience. 
(b) Performance of the proposed LSDT-Large model.}
\label{fig.internal_confusion}
\end{figure}

For clinical reference on the internal four-class cohort, we additionally report the human-expert and MLLM evaluations following the protocol above. Table~\ref{tab:internal_main} reports the three-reader average as the human-expert benchmark, and Fig.~\ref{fig.internal_confusion} visualizes the senior-reader confusion pattern using the class labels Normal, BPH, nsPCa, and sPCa.

\begin{table}[t!]
\small
\caption{Internal PCa-HSD four-class comparison (ACC and JointRecall)}
\label{tab:internal_main}
\centering
\begin{tabular}{lcc}
\hline
Method & ACC & JointRecall \\
\hline 
Human Experts & $0.584\pm0.039$ & $0.696\pm0.020$ \\
\hline
ResNet-50~\cite{he2016deep} & $0.526\pm0.049$ & $0.672\pm0.047$ \\
Swin-T~\cite{liu2021swin} & $0.561\pm0.047$ & $0.733\pm0.081$ \\
ViT-Base~\cite{dosovitskiy2021image} & $0.587\pm0.054$ & $0.753\pm0.057$ \\
ViT-Large~\cite{dosovitskiy2021image} & $0.596\pm0.035$ & $0.698\pm0.048$ \\
BiomedCLIP-Vision~\cite{zhang2023biomedclip} & $0.604\pm0.073$ & $0.753\pm0.024$ \\
Omnirad-Vision~\cite{zedda2025omnirad} & $0.596\pm0.053$ &  $0.770\pm0.067$ \\
\hline
LSDT-ResNet50 & $0.567\pm0.061$ & $0.674\pm0.134$ \\
LSDT-Swin-T & $0.616\pm0.061$ & $0.744\pm0.063$ \\
LSDT-Base & $0.616\pm0.059$ & $0.791\pm0.041$ \\
LSDT-Large & $\bm{0.633\pm0.050}$ & $\bm{0.768\pm0.040}$ \\
LSDT-BiomedCLIP & $0.625\pm0.052$ & $0.767\pm0.024$ \\
LSDT-Omnirad & $0.604\pm0.057$ & $0.765\pm0.029$ \\
\hline
\end{tabular}
\end{table}

\section{Discussions}

\subsection{Results on Internal PCa-HSD Four-Class Cohort (Primary Endpoint)}
As shown in Table~\ref{tab:internal_main}, conventional foundation model architectures achieve only moderate performance on the proposed four-class task, with ResNet-50 ($0.526 \pm 0.049$), Swin-T ($0.561 \pm 0.047$), ViT-Base ($0.587 \pm 0.054$), and ViT-Large ($0.596 \pm 0.035$) in terms of accuracy. We further evaluate foundation models that undergo additional post-training on large-scale medical data, including BiomedCLIP-Vision ($0.604 \pm 0.073$) and OmniRad-Vision ($0.596 \pm 0.053$), both of which exhibit only marginal improvements over general-domain counterparts. These results indicate that current foundation models, whether general-purpose or medical-domain, remain insufficient for fine-grained prostate mpMRI classification. This observation highlights the intrinsic difficulty of the task.

Several factors contribute to this limitation. Prostate MRI datasets are typically small-scale and exhibit substantial inter-patient variability, which restricts the generalization ability of large-scale pre-trained models. Furthermore, the prostate gland varies significantly in position, size, and shape across patients, introducing additional spatial heterogeneity that is not explicitly modeled by standard architectures. In addition, the presence of extensive background and surrounding tissues introduces irrelevant information, diluting the discriminative signal of the target gland. More importantly, the four-class task itself is intrinsically difficult due to the strong overlap in mpMRI appearance among different tissue types. In particular, BPH can exhibit imaging characteristics highly similar to prostate cancer, with overlapping ADC and T2-weighted signals and even mimicking malignant lesions. Consequently, learning robust and discriminative representations for fine-grained prostate disease classification remains a nontrivial task even for advanced deep learning models. 

Despite these challenges, LSDT achieves consistent performance gains, reaching the best accuracy of $\mathbf{0.633 \pm 0.050}$ and JointRecall of $\mathbf{0.768 \pm 0.040}$. For clinical reference, the radiologists with 15, 5, and 3 years of prostate MRI experience achieved ACC/JointRecall of $0.622/0.718$, $0.587/0.701$, and $0.544/0.669$, respectively, corresponding to the three-reader average reported in Table~\ref{tab:internal_main}. Notably, LSDT-Large achieves the highest ACC and JointRecall among all methods and human readers---surpassing even the 15-year senior radiologist in overall accuracy and exceeding all three readers in cancer detection sensitivity (JointRecall). This suggests that the proposed model reaches an expert-comparable operating level on the internal cohort, supporting its potential clinical relevance. Importantly, the proposed paradigm leads to consistent performance improvements across different backbone architectures. Among general-purpose vision backbones, ResNet-50 improves from $0.526\pm0.049$ to $0.567\pm0.061$, Swin-T from $0.561\pm0.047$ to $0.616\pm0.061$, and ViT-Base from $0.587\pm0.054$ to $0.616\pm0.059$. The most notable improvement is observed in ViT-Large, where the accuracy increases from $0.596\pm0.035$ to $0.633\pm0.050$, along with a JointRecall improvement from $0.698\pm0.048$ to $0.768\pm0.040$. For medical-domain foundation models, LSDT-BiomedCLIP improves from $0.604\pm0.073$ to $0.625\pm0.052$ and LSDT-OmniRad from $0.596\pm0.053$ to $0.604\pm0.057$, demonstrating that our framework consistently enhances performance irrespective of the pre-training domain. These results indicate that the proposed SAM3-based paradigm is not limited to a specific architecture, but instead provides a general and effective strategy for enhancing prostate MRI classification. By leveraging the zero-shot capability of SAM3, our method generates prostate-specific masks via text prompts, enabling precise localization of the gland without requiring manual annotations. This region-aware mechanism significantly reduces background noise and constrains the learning process to clinically relevant regions, alleviating spatial variability and improving generalization under limited data conditions. Furthermore, the proposed slice-wise feature interaction module facilitates effective information exchange across slices, allowing the model to better capture slice sequential contextual dependencies within multi-modal MRI. 

As shown in Fig.~\ref{fig.internal_confusion}, compared with the senior radiologist benchmark, LSDT-Large more effectively separates BPH from cancer and reduces confusion between nsPCa and sPCa, while both reader and model remain challenged by nsPCa. Clinically, this suggests potential value for reducing over-calling of benign lesions and improving risk stratification consistency.

\begin{table}[!t]
\small
\caption{Comparison with MLLMs on the internal PCa-HSD four-class cohort}
\label{tab:internal_vlm}
\centering
\begin{tabular}{lcc}
\hline
Method & ACC & JointRecall \\
\hline
Claude Sonnet 4.5 & $0.299$ & $0.515$ \\
GPT 5mini & $0.299$ & $0.515$ \\
GPT 5.2 & $0.314$ & $0.486$ \\
Qwen 3.6 Plus & $0.334$ & $0.512$ \\
Gemini 3.1 Pro Preview & $0.344$ & $0.525$ \\
Gemini 3 Flash Preview & $0.375$ & $0.584$ \\
\hline
LSDT-Large & $\bm{0.633\pm0.050}$ & $\bm{0.768\pm0.040}$ \\
\hline
\end{tabular}
\end{table}

To further position the proposed method against recent multimodal foundation models, we additionally compare LSDT-Large with GPT-5 series, Gemini 3 series, Claude Sonnet 4.5, and Qwen 3.6 Plus MLLM baselines on the same internal PCa-HSD cohort. Since these MLLMs follow a prompt-based inference paradigm rather than supervised training on PCa-HSD, we report them separately in Table~\ref{tab:internal_vlm}. This comparison is intended to assess whether off-the-shelf MLLM reasoning is sufficient for fine-grained prostate MRI stratification under the four-class setting. A likely reason is that this task requires task-specific knowledge, prostate localization, cross-slice information aggregation, and pathology-grounded supervision, which are not adequately captured by generic prompt-based multimodal reasoning alone.

\subsection{Results on Public PI-CAI binary Cohort (External Validation)}

\begin{figure}[!t]
\centering
\includegraphics[width=\linewidth]{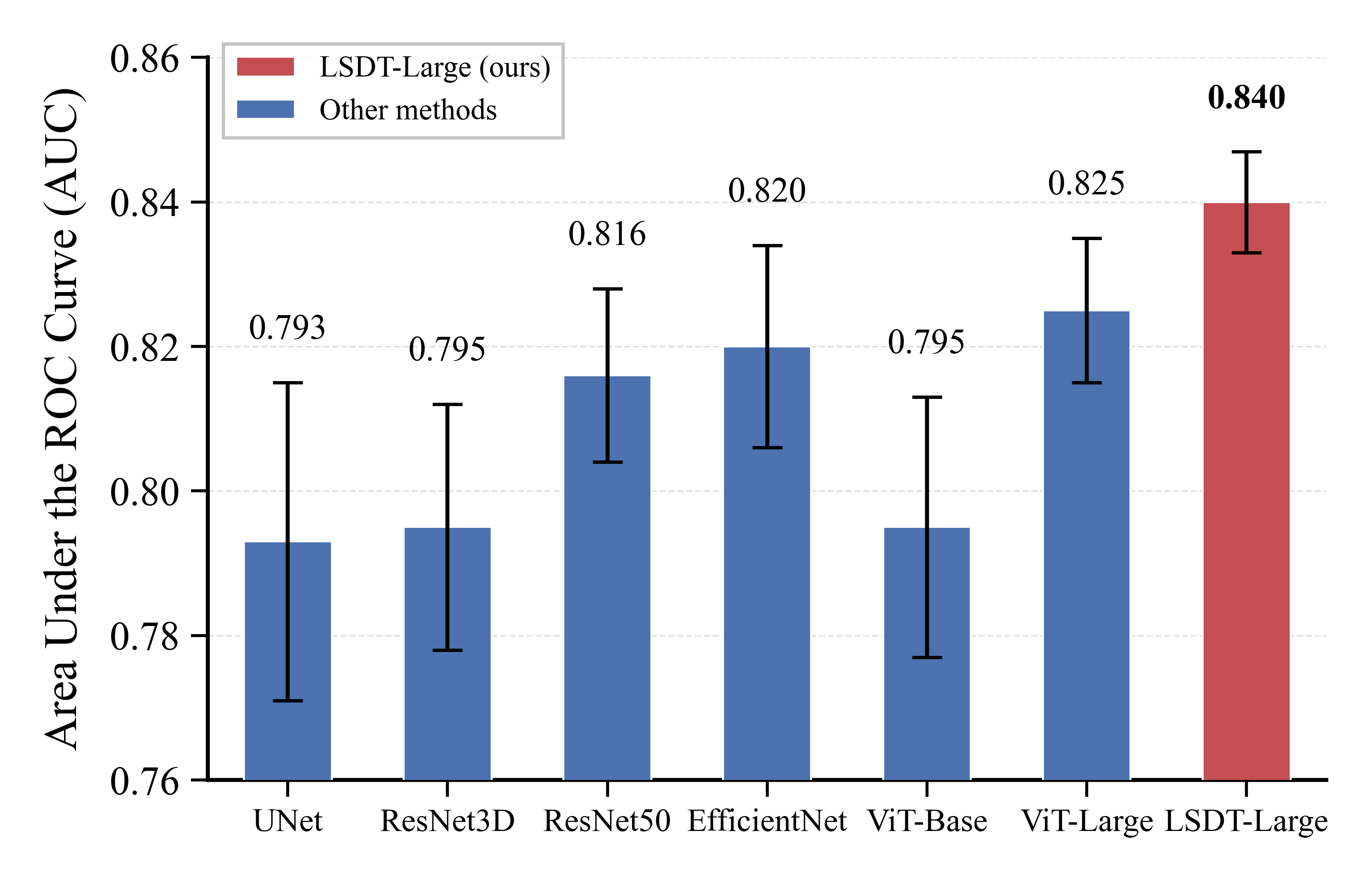}
\caption{Public binary-cohort comparison (AUC) on the PI-CAI dataset. LSDT-Large achieves the highest AUC of $0.840 \pm 0.007$, demonstrating strong cross-institutional generalization.}
\label{fig.external_auc}
\end{figure}

To further assess generalization and mitigate potential institutional bias, we evaluate our methods on the public PI-CAI binary cohort under adapted official implementations and recommended training protocols. 

As shown in Fig.~\ref{fig.external_auc}, among medically common baselines, UNet~\cite{ronneberger2015u} and ResNet3D~\cite{hara2018can} achieve moderate performance ($0.793 \pm 0.022$ and $0.795 \pm 0.017$, respectively), while EfficientNet~\cite{tan2019efficientnet} obtains $0.820 \pm 0.014$. For general vision foundation models, a clear scaling trend is observed (ResNet50~\cite{he2016deep} $0.816 \pm 0.012$, ViT-Base~\cite{dosovitskiy2021image} $0.795 \pm 0.018$, ViT-Large~\cite{dosovitskiy2021image} $0.825 \pm 0.010$), confirming that large-scale pre-training on natural images offers an advantage over task-specific architectural designs alone for this task.

LSDT-Large achieves the highest overall AUC of $\mathbf{0.840 \pm 0.007}$. Notably, on the PI-CAI dataset, since the prostate gland has been largely cropped to the central region during preprocessing, we adapt LSDT-Large by incorporating the SAM3-generated mask as an additional input channel (rather than hard masking) to provide weak guidance, and use learnable parameters for slice-level feature fusion. The SAM3-guided masking mechanism suppresses confounding background signals from surrounding pelvic tissues and focuses representational capacity on the prostate region at zero annotation cost, a benefit that is especially valuable on multi-center data where anatomical variability is amplified. Moreover, the slice-level attention fusion module learns to adaptively prioritize diagnostically informative slices while down-weighting uninformative ones, enabling effective aggregation of volumetric lesion context without requiring explicit slice selection or manual priors.

Critically, LSDT-Large achieves consistent superiority across the internal four-class cohort (ACC $0.633$, JointRecall $0.768$) and the external binary cohort (AUC $0.840$). The internal task demands fine-grained discrimination across four histopathological categories with acquisition characteristics from two institutions, whereas the external PI-CAI cohort introduces substantial cross-center variability in imaging parameters and patient demographics. The ability of the same model to excel under both settings indicates that our proposed framework is not dataset-specific optimizations but rather general-purpose mechanisms that robustly enhance prostate mpMRI analysis across tasks and institutions.

\begin{figure}[!t]
\centering
\captionsetup[subfigure]{labelformat=empty}
\begin{subfigure}[b]{0.42\linewidth}
    \includegraphics[width=\linewidth]{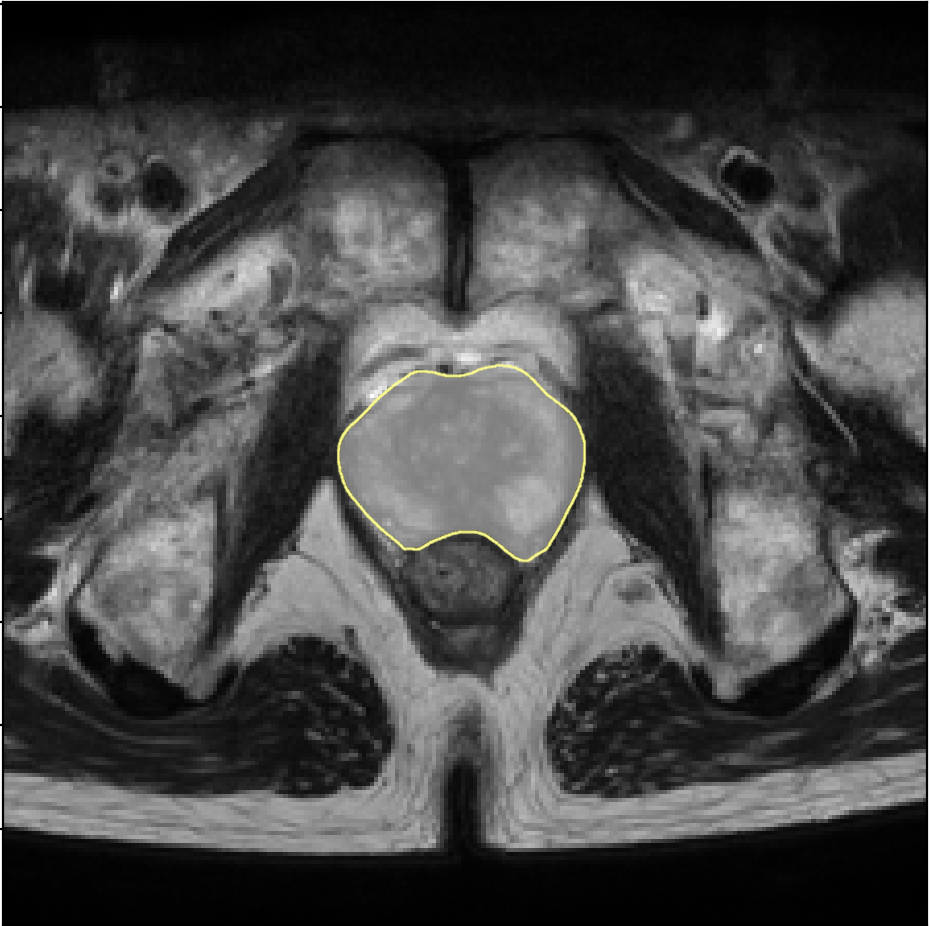}
    \caption{\mbox{(a) Ground truth}}
\end{subfigure}%
\hfill
\begin{subfigure}[b]{0.42\linewidth}
    \includegraphics[width=\linewidth]{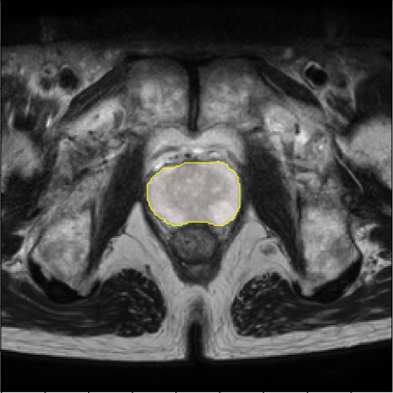}
    \caption{\mbox{(b) SAM3 language-guided}}
\end{subfigure}%
\\[0.5em]
\begin{subfigure}[b]{0.42\linewidth}
    \includegraphics[width=\linewidth]{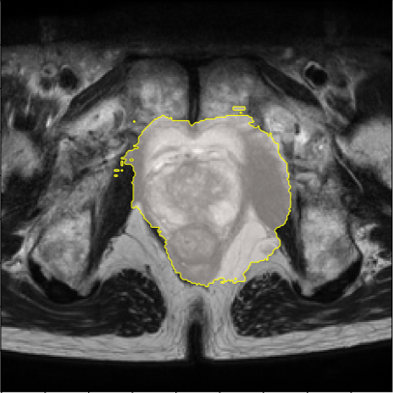}
    \caption{\mbox{(c) MedSAM 1/2-window}}
\end{subfigure}%
\hfill
\begin{subfigure}[b]{0.42\linewidth}
    \includegraphics[width=\linewidth]{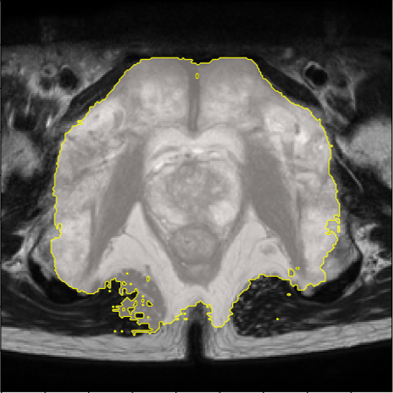}
    \caption{\mbox{(d) MedSAM 3/4-window}}
\end{subfigure}
\caption{
Prostate segmentation masks of different strategies.
(a) Ground-truth annotation. 
(b) SAM3 with the text prompt ``prostate'';
(c) MedSAM with a 1/2-window bounding box prompt;
(d) MedSAM with a 3/4-window bounding box prompt.
}
\label{fig.mask_comparison}
\end{figure}

\subsection{Ablation Study}
\subsubsection{Ablation Study 1: Effect of Mask Generation Strategies}

We compare two representative promptable segmentation models with different prompting paradigms. MedSAM is a medical-domain adaptation of the Segment Anything Model that relies on geometric prompts such as bounding boxes, whereas SAM3 is a vision foundation model capable of open-vocabulary segmentation via text prompts.

As shown in Table~\ref{tab:ablation1}, the mask generation strategy has a consistent impact on classification performance across different backbone architectures. For both ResNet-50 and ViT-Large, SAM3 achieves the best performance, followed by MedSAM with larger prompt windows (3/4) and smaller windows (1/2). For example, in the ViT-Large setting, accuracy improves from $0.613\pm0.042$ (MedSAM 3/4) and $0.619\pm0.049$ (MedSAM 1/2) to $\mathbf{0.633\pm0.050}$ with SAM3. This trend is consistent with the qualitative results in Fig.~\ref{fig.mask_comparison}. SAM3 produces masks that are more anatomically accurate and closer to the ground truth, whereas MedSAM is sensitive to bounding box design: smaller boxes may miss parts of the prostate, while larger boxes introduce background noise. Such differences in mask quality have a direct impact on downstream classification. More accurate localization enables the model to focus on prostate-specific features and reduces interference from irrelevant regions, leading to improved discriminative representations. Consequently, SAM3 consistently outperforms MedSAM across different configurations.

These results demonstrate that mask quality is a key factor in prostate MRI classification, and that improved anatomical localization is essential for enhancing fine-grained classification performance.

\begin{table}[t!]
\footnotesize
\caption{Ablation Study I: Effect of Mask Generation Strategy on PCa-HSD (ACC and JointRecall)}
\label{tab:ablation1}
\centering
\setlength{\tabcolsep}{2pt}
\begin{tabular}{lccc}
\hline
Configuration & Mask Strategy & ACC & JointRecall \\
\hline
LSDT-ResNet50 & MedSAM(3/4) & $0.555\pm0.030$ & $0.735\pm0.084$ \\
LSDT-ResNet50 & MedSAM(1/2) & $0.570\pm0.040$ & $0.753\pm0.060$ \\
LSDT-ResNet50 & SAM3 &$0.567\pm0.061$ & $0.674\pm0.134$ \\
LSDT-Large & MedSAM(3/4) & $0.613\pm0.042$ & $0.704\pm0.090$ \\
LSDT-Large & MedSAM(1/2) & $0.619\pm0.049$ & $0.765\pm0.050$ \\
LSDT-Large & SAM3 &$\bm{0.633\pm0.050}$ & $\bm{0.768\pm0.040}$ \\
\hline
\end{tabular}
\end{table}

\subsubsection{Ablation Study 2: Slice Fusion Strategy}
We further investigate the impact of different slice fusion strategies on downstream classification performance. In prostate MRI, the gland typically occupies only a limited subset of slices (approximately one-quarter to one-half of the full volume in PCa-HSD), while the remaining slices contain little or no relevant information. In addition, the prostate exhibits substantial inter-patient variability in its slice location and volumetric extent. These characteristics make it essential to effectively aggregate gland-specific information and model cross-slice interactions.

As shown in Table~\ref{tab:ablation2}, incorporating slice fusion consistently improves accuracy compared to the baseline without explicit fusion. Among different strategies, attention-based fusion achieves the best performance, reaching $\mathbf{0.633\pm0.050}$ in accuracy and $\mathbf{0.768\pm0.040}$ in JointRecall. In comparison, mean pooling ($0.625\pm0.064$), concatenation ($0.619\pm0.041$), and CNN-based fusion ($0.628\pm0.054$) yield inferior results.

These differences can be attributed to the limitations of each fusion strategy. Mean pooling assumes equal contribution from all slices and may dilute discriminative features from informative regions. Concatenation increases feature dimensionality but does not explicitly model inter-slice relationships. CNN-based fusion captures local dependencies along the slice dimension but remains limited in modeling long-range interactions across distant slices.

In contrast, the attention mechanism enables adaptive weighting of slice features, allowing the model to emphasize slices containing relevant prostate regions while suppressing irrelevant ones. Moreover, attention allows each slice to interact with all others, effectively capturing global dependencies within the sequence and modeling long-range contextual relationships. This property is particularly beneficial in prostate MRI, where informative signals are sparsely distributed across slices and require global aggregation.

These results demonstrate that effective cross-slice interaction is essential for multi-slice prostate MRI analysis, and that attention-based fusion provides a more suitable solution for capturing slice sequential contextual information in fine-grained classification tasks.

\begin{table}[!t]
\footnotesize
\caption{Ablation Study II: Effect of Slice Fusion Method on PCa-HSD (ACC and JointRecall)}
\label{tab:ablation2}
\centering
\begin{tabular}{lccc}
\hline
Configuration & Fusion method & ACC & JointRecall \\
\hline
LSDT-Large & Sum & $0.610\pm0.060$ & $0.779\pm0.023$ \\
LSDT-Large & Concat & $0.619\pm0.041$ & $0.718\pm0.021$\\
LSDT-Large & Mean & $0.625\pm0.064$ & $0.788\pm0.058$ \\
LSDT-Large & CNN & $0.628\pm0.054$ & $0.785\pm0.036$ \\
LSDT-Large & Att & $\bm{0.633\pm0.050}$ & $\bm{0.768\pm0.040}$ \\
\hline
\end{tabular}
\end{table}

\subsection{Limitations and Future Work}
This study has several limitations. The proposed model is evaluated on a dataset collected from two institutions within the same city, which may limit its generalizability across different scanners, acquisition protocols, and patient populations. The current framework does not fully exploit the three-dimensional nature of MRI data. Although slice-wise fusion captures inter-slice dependencies, volumetric context remains underutilized, as 3D models have shown advantages in modeling voxel-level spatial information~\cite{zhang2020inter}. Meanwhile, SAM3 enables automatic ROI extraction through zero-shot segmentation. It is not specifically optimized for prostate MRI and may underperform in challenging regions with low contrast or ambiguous anatomy, particularly within the peripheral zone, thereby affecting cancer detection performance~\cite{bura2021mri,gadalla2023diagnostic}. This limitation can introduce noise into the extracted regions of interest and consequently affect downstream classification performance.

Finally, we emphasize that fine-grained prostate cancer classification based on MRI is inherently challenging and serves as the primary motivation of this work. The current framework relies solely on imaging data, whereas incorporating quantitative clinical measurements, such as prostate-specific antigen (PSA) may further improve diagnostic performance through multimodal fusion~\cite{deforche2024improved}. In addition, model interpretability is a critical requirement for clinical deployment. Recent studies combining anatomical priors with explainable AI frameworks have shown promising results in improving the reliability of risk assessment, providing a valuable direction for further research~\cite{ahmed2026explainable,yu2024explainability}. Future work may incorporate interpretability techniques to provide more transparent and clinically consistent explanations of model predictions.

\section{Conclusion}

This work introduces a pathology-grounded four-class classification task for prostate mpMRI, enabling clinically meaningful differentiation among normal tissue, benign hyperplasia, non-significant cancer, and significant cancer. To address the challenges of fine-grained classification and multi-modal data, we propose a segmentation-assisted framework that incorporates prostate-aware masking and slice-level feature fusion. Experimental results show that the proposed method consistently improves accuracy by reducing background interference, enhancing anatomical localization, and enabling cross-slice interaction. The best model achieves an accuracy of $\mathbf{0.633 \pm 0.050}$ and JointRecall of $\mathbf{0.768 \pm 0.040}$ on 344 patients.

Overall, this work highlights the importance of pathology-grounded supervision and anatomical priors for fine-grained prostate cancer classification, and provides a practical step toward more reliable and clinically actionable MRI-based diagnosis.

\bibliographystyle{unsrtnat}
\bibliography{IEEEabrv,reference}

\end{document}